# Investigation of Multimodal Features, Classifiers and Fusion Methods for Emotion Recognition

Zheng Lian
National Laboratory of Pattern Recognition, Institute of Automation Chinese Academy of Sciences, Beijing, China
lianzheng2016@ia.ac.cn

Ya Li
National Laboratory of Pattern Recognition, Institute of Automation Chinese Academy of Sciences, Beijing, China
yli@nlpr.ia.ac.cn

Jianhua Tao
National Laboratory of Pattern Recognition, CAS Center for Excellence in Brain Science and Intelligence Technology, Institute of Automation Chinese Academy of Sciences, Beijing, China
jhtao@nlpr.ia.ac.cn

Jian Huang
National Laboratory of Pattern Recognition, Institute of Automation Chinese Academy of Sciences, Beijing, China
jian.huang@nlpr.ia.ac.cn

## ABSTRACT

Automatic emotion recognition is a challenging task. In this paper, we present our effort for the audio-video based sub-challenge of the Emotion Recognition in the Wild (EmotiW) 2018 challenge, which requires participants to assign a single emotion label to the video clip from the six universal emotions (Anger, Disgust, Fear, Happiness, Sad and Surprise) and Neutral. The proposed multimodal emotion recognition system takes into account audio, video and text information. Besides handcraft features, we also extract bottleneck features from deep neural networks (DNNs) via transfer learning. Both temporal classifiers and non-temporal classifiers are evaluated to obtain the best unimodal emotion classification result. Then emotion possibilities are calculated and fused by the Beam Search Fusion (BS-Fusion). We test our method[1] in the EmotiW 2018 challenge and we a gain promising result: 60.34% on the testing dataset. Compared with the baseline system, there is a significant improvement. What's more, our result is only 1.5% lower than the winner's.

## 1 INTRODUCTION

With the development of artificial intelligence, there is an explosion of interest in realizing more natural human-machine interaction (HMI) systems. The emotion, as an important aspect of HMI, is also attracting more and more attention. Due to the complexity of emotion recognition and the diversity of application scenarios, the single modality is difficult to meet the demand. Multimodal recognition methods, which take into account the audio, video, text and biological information, can improve the recognition performance.

The audio-video based sub-challenge of the Emotion Recognition in the Wild (EmotiW) challenge plays an important role in the emotion recognition. The Acted Facial Expressions in the Wild (AFEW) dataset [1] is the dataset of the EmotiW challenge. Organizers provide an open platform for participators to evaluate their recognition systems. The first EmotiW challenge was organized in 2013. This year is the 6[th] challenge. The recognition accuracy on the seven emotions increases every year: 41.03%[2], 50.37% [3], 53.80%[4], 59.02% [5], 60.34%[6].

It is important to extract more discriminative features in the emotion classification. Before the popularity of deep neural networks (DNNs), frame-level handcraft features are wildly studied and utilized [2, 7, 8], including Histogram of Oriented Gradient (HOG) [9], Local Binary Patterns (LBP) [10], Local Phase Quantization (LPQ) [11] and Scale Invariant Feature Transform (SIFT) [12]. Three Orthogonal Planes (TOP) [13], summarizing functionals (FUN), Fisher Vector encoding (FV) [14], Spatial Pyramid Matching (SPM) [15] and Bag of Words (BOW) are also utilized to capture temporal information [7, 16]. Now the DNNs based approach generates the state-of-the-art performance in many fields [17-21]. However, due to limited training samples in the AFEW database, complex DNNs are difficult to train [22]. To deal with that problem, transfer learning is adopted. Then bottleneck features are extracted from fine-tuned models [5, 23, 24].

The classifiers are also important in the emotion recognition. Liu et al. [3] exploited Partial Least Squares (PLS), Logistic Regression (LR) and Kernel Support Vector Machine (KSVM) operating in vector space to classify data points on Riemannian manifolds for emotion recognition. Kaya et al. [25] chose Extreme Learning Machines (ELM) and Kernel Extreme Learning

---
[1]NLPR's method: https://github.com/zeroQiaoba/EmotiW2018

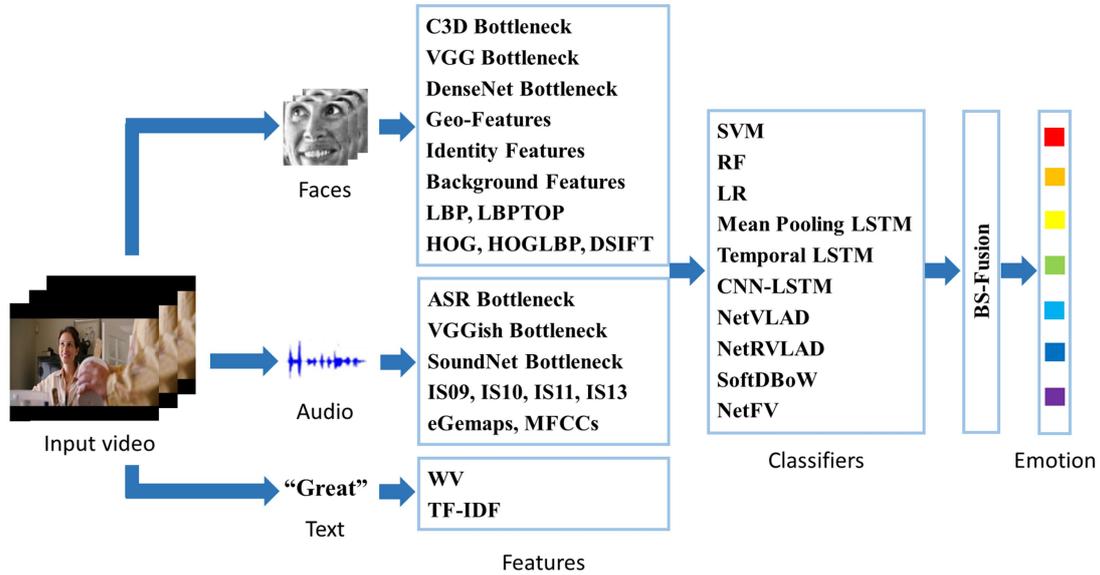

**Figure 1: An overview of the proposed multimodal emotion recognition system. Features from different modalities are trained individually based on multiple classifiers. Emotion possibilities are fused by the BS-Fusion.**

Machines (KELM) for modeling modality-special features, which were faster and more accurate than SVM. Recently, many temporal models are also tested, such as Long Short-Term Memory (LSTM) [26], Gated Recurrent Unit (GRU) [27] and 3D Convolution Networks (C3D) [28].

To gain better performance, fusion methods that merge different modalities are essential. Fusion methods can be classified into feature level fusion (or called early fusion), decision level fusion (or called late fusion) and model level fusion. Most teams chose late fusion in pervious challenges [5, 23, 29]. Vielzeuf et al. [30] discussed five fusion methods: Majority Vote, Mean, ModDrop, Score Tree and Weighted Mean. They found that Weighted Mean was the most effective fusion method, which had less risk of overfitting. Ouyang et al. [22] utilized reinforcement learning strategy to find the best fusion weight.

In EmotiW 2018 [31], we participate in the Audio-Video based sub-challenge. The task is to assign a single emotion label to the video clip and classification accuracy is the comparison metric. In this paper, we propose our multimodal emotion recognition system, which is shown in Fig. 1. Features from different modalities are trained individually based on multiple classifiers. Emotion possibilities are fused by the BS-Fusion. Compared with the previous solutions in EmotiW challenges, our innovations mainly focus on three parts:

1. Multimodal features: To our best knowledge, it is the first time to take into account text, identity and background information.
2. Classifiers: Different types of aggregation models are investigated, including NetFV, NetVLAD, NetRVLAD and SoftDBoW[32].
3. Fusion methods: The Beam Search Fusion (BS-Fusion) is proposed for the modal selection and weight determination.

The rest of paper is organized as follow. Multimodal features and various classifiers are illustrated in Section 2 and Section 3, respectively. In Section 4, we focus on our proposed BS-Fusion. Datasets and experimental results are illustrated in Section 5 and Section 6, respectively. Section 7 concludes the whole paper.

## 2 MULTIMODAL FEATURES

In our approach, audio, video and text features are taken into account to improve the recognition performance. Besides handcraft features, bottleneck features extracted from fine-tuned models are also considered.

### 2.1 Audio Features

In this section, multiple audio features are discussed. Besides handcraft feature sets, bottleneck features of the automatic speech recognition (ASR) acoustic model, SoundNet and VGGish are also evaluated.

*2.1.1 OpenSMILE-based Audio Features.* The OpenSMILE toolkit [33] is utilized to extract audio feature sets, including eGemaps (eGeMAPSv-01.conf) [34], IS09 (IS09_emotion.conf), IS10 (IS10_paraling.conf), IS11 (IS11_speaker_state.conf), IS13 (IS13_ComParE.conf) and MFCC (MFCC12_0_D_A.conf).

To extract those feature sets, the acoustic low-level descriptors (LLDs), covering spectral, cepstral, prosodic and voice quality information, are first extracted within a 25ms frame with a window shift of 10ms. Then statistical functions such as mean and maximum are calculated over LLDs to get segment-level features. We test two segment lengths in the paper: 100ms and the length of the whole utterance.

*2.1.2 ASR Bottleneck Features.* We extract bottleneck features from the ASR acoustic model. At first, we train a Chinese ASR

system with the 500 hours spontaneous and accented Mandarin speech corpus. The ASR acoustic model has six hidden layers. The first five layers have 1024 nodes and the last layer has 60 nodes. As most speakers are spoken in English in the AFEW dataset, we fine-tune the Chinese ASR system with the 300 hours English speech corpus due to the limited English corpus. Then, we extract bottleneck features from two acoustic models: the English ASR acoustic model and the Chinese ASR acoustic model.

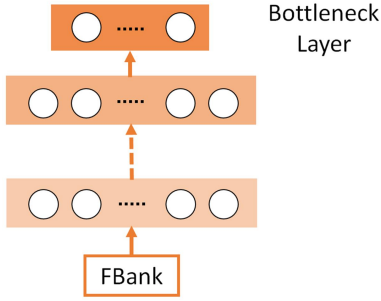

**Figure 2: The architecture of our ASR acoustic model. FBank features extracted from waveforms are used as inputs. The last layer of the ASR acoustic model is treated as the bottleneck layer.**

*2.1.3 SoundNet Bottleneck Features.* We extract bottleneck features from the SoundNet network [35], which learns rich natural sound representation by capitalizing on large amounts of unlabeled sound data collected in the wild. The SoundNet network is a 1-dimensional convolutional network, which consists of full convolutional layers and pooling layers.

In this paper, we divide raw waveforms into multiple 1s segments. Then those segments are treated as inputs to the network and we extract SoundNet Bottleneck Features from the *conv7* layer in Fig.3.

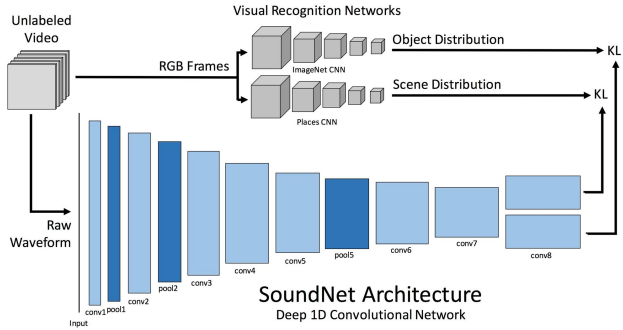

**Figure 3: The architecture of the SoundNet network [35]. Visual knowledge is transferred into the sound modality using unlabeled video as a bridge.**

*2.1.4 VGGish Bottleneck Features.* The VGGish network [36] is trained on AudioSet [37], which contains over 2 million human-labeled 10s YouTube video soundtracks with more than 600 audio event classes.

In this paper, the VGGish network is used as the feature extractor. We divide raw waveforms into multiple 1s segments. Log spectrograms extracted from segments are treated as inputs. VGGish extracts semantically meaningful, high-level 128D embedding features from *fc2*. Then Principal Component Analysis (PCA) is utilized to extract normalized features.

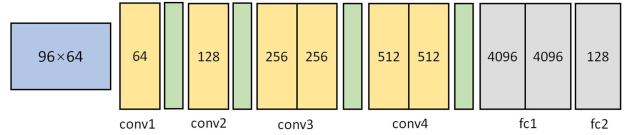

**Figure 4: The structure of the VGGish network. The input log spectrogram is 96×64. Yellow boxes, green boxes and grey boxes denote the 2D convolutional layers, max pooling layers and fully-connected layers, respectively. The number inside of the yellow box is the number of filters and the number inside of the grey box is the number of neurons.**

## 2.2 Video Features

In this paper, we extract multiple video features. Besides handcraft features such as Local Binary Patterns from Three Orthogonal Planes (LBPTOP) [13], HOG and Dense SIFT (DSIFT), bottleneck features extracted from VGG, DenseNet and C3D are also considered. Furthermore, we take into account geometry features, background features and identity features.

*2.2.1 Handcraft Video Features.* In general, facial features consist of two parts: appearance features and geometry features.

As for appearance features, LBPTOP features are wildly used in previous EmotiW challenges. Basic LBP features have 59 features while using uniform code. LBPTOP features extend LBP from two dimensions to three dimensions, which apply the relevant descriptor on XY, XT and YT planes independently and concatenate histograms together.

Besides LBPTOP features, LBP, HOG, HOGLBP and DSIFT features are also tested. HOGLBP features apply the HOG descriptor on the XY plane, LBP descriptor on the XT and YT plane and then concatenate them together. As for DSIFT features, it is equivalent to performing the SIFT descriptor on a dense grid of locations on an image at a fixed scale and orientation.

As for geometry features, the head pose and landmarks are considered. Emotion is related with landmarks and the head pose. When people feel *neutral*, movement of landmarks is relatively small. When people feel *sad*, they tend to lower their heads. Therefore, we take into account those features, which are marked as Geo-Features.

*2.2.2 CNN Bottleneck Features.* To extract bottleneck features from images, the VGG (configuration "D") [38] and DenseNet-BC [39] structure are chosen.

In this paper, the VGG and DenseNet-BC network are pre-trained on ImageNet [40] and fine-tuned on the (Facial Expression Recognition +) FER+ [41] and Static Facial Expression in the Wild (SFEW) 2.0 database [42]. Grey-scale images are treated as inputs. As for the VGG network, we extract bottleneck features from *conv5-b*, *conv5-c*, *fc1* and *fc2* in Fig.5. As for the DenseNet-BC structure, we extract bottleneck features from the last mean pooling layer, which is marked as *pool3*.

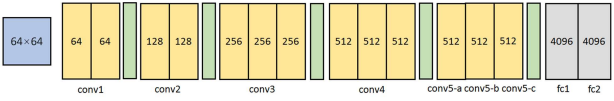

**Figure 5: The structure of the VGG network. The input image is 64×64 pixels. Meanings of other components are the same as definitions in Fig. 4.**

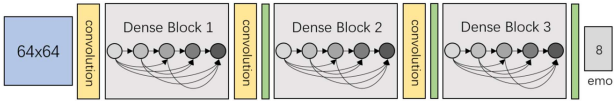

**Figure 6: The structure of the DenseNet-BC [43] network. The input image is 64×64 pixels. There are three Dense Block. Yellow boxes and green boxes denote the convolutional layers and the mean pooling layers, respectively.**

*2.2.3 C3D Features.* The C3D network is an extension of the 2D convolutional process, which captures spatial-temporal features from videos. The C3D network shows its performance in previous EmotiW challenges [5, 22]. The architecture of C3D is shown in Fig. 7.

In this paper, C3D network is pre-trained on sports1M [44] and fine-tuned on the AFEW database. It takes continuous 16 frames as inputs with 8 overlapped frames. Outputs of *fc6* are exploited as bottleneck features.

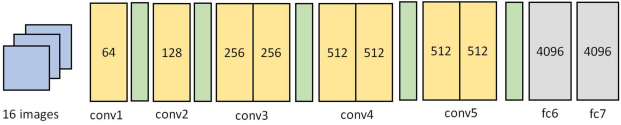

**Figure 7: The structure of the C3D network. It takes continuous 16 images as inputs. Each image is cropped into 112×112 pixels. Yellow boxes denote 3D convolutional layers. Meanings of other components are the same as the definitions in Fig. 4.**

*2.2.4 Background Features.* Background information is helpful to judge emotion states. *Fear* is often accompanied with a dim environment. *Happy* is often accompanied with a bright environment. To take into account background information, we take the Inception network [45] as the feature extractor, which is pre-trained on ImageNet. Original frames extracted from videos are passed into the network. The last mean pooling layer is treated as the bottleneck layer. Then PCA is utilized to extract normalized features and reduce feature dimensions.

*2.2.5 Identity Features.* Identity information also counts. As some samples in the AFEW database are continuous, their emotions have high possibilities to be continuous as well.

In the experiment, SeetaFace[1] is utilized to extract identity features. SeetaFace identification is based on deep convolutional neural network (DCNN). Specifically, it is an implementation of the VIPLFaceNet [46], which consists of 7 convolutional layers and 2 fully-connected layers with the input size of 256x256x3. In the SeetaFace open-source face identification toolkit, outputs of 2048 nodes of the *FC2* layer in the VIPLFaceNet are exploited as the feature of the input face.

## 2.3 Text Features

Contents in audios reflect the emotion. For example, dirty words such as 'fuck' and 'shit' are common when people are angry. 'Sorry' is always utilized to express one's guilt about others. People often use 'oh my god' to express their surprise.

To take into account text information, Term Frequency–Inverse Document Frequency (TF-IDF) [47] and Word Vectors (WV) [48] are utilized to extract computable features from raw texts.

*2.3.1 TF-IDF.* TF-IDF is a numerical statistic that is intended to reflect how important a word is to a document. TF means term frequency while IDF means inverse document frequency. The TF-IDF value increases proportionally to the number of times a word appears in the document and is offset by the frequency of the word in the corpus.

$$TF - IDF(t,d) = TF(t,d) \times IDF(t) \quad (1)$$

$$IDF(t) = \log \frac{1 + n_d}{1 + df(d,t)} + 1 \quad (2)$$

where *TF(t, d)* shows the number of times the word *t* appears in the document *d*. $n_d$ is the total number of documents and $df(d, t)$ is the number of documents that contain the word *t*.

*2.3.2 WV.* Word vectors are the high-level representation of words, which learn grammatical relations between words through a large corpus.

In the paper, we utilize pre-trained FastText word vectors. It has 1 million word vectors trained on the Wikipedia 2017, UMBC webbase corpus and statmt.org news dataset. Each word can be mapped into 300-D computable vectors.

## 3 CLASSIFIERS

Besides classic classifiers such as SVM, Random Forest (RF) and LR, we also test temporal models, including Mean Pooling LSTM, temporal LSTM and CNN-LSTM Model. Furthermore, several types of aggregation models are also investigated: NetVLAD, NetRVLAD, SoftDBoW and NetFV.

## 3.1 Mean Pooling LSTM

As for Mean Pooling LSTM, we use a one-layer LSTM and average the time-step outputs as the video representation in the encoder and a fully-connected layer in the decoder. The softmax layer is treated as the classifier. The structure of Mean Pooling LSTM is shown in Fig. 8.

---
[1]SeetaFace: https://github.com/seetaface/SeetaFaceEngine

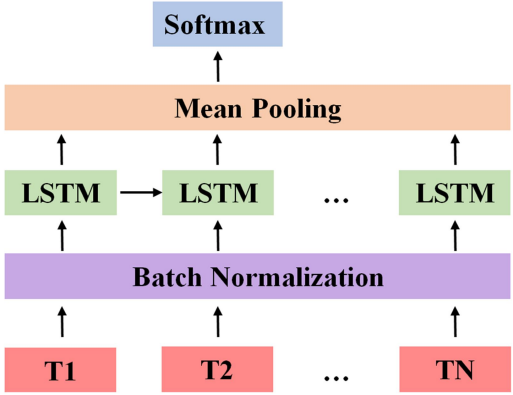

**Figure 8: The structure of Mean Pooling LSTM. Red boxes represent features in different time steps.**

### 3.2 Temporal LSTM

To consider more contextual information, we propose Temporal LSTM. The difference between Temporal LSTM and Mean Pooling LSTM mainly focuses on inputs. Instead of processing on one time step features, features in the same window are concatenate together as inputs in Temporal LSTM. The overlap size can be adjusted. If the overlap size is set to be 0, adjacent windows are processed independently. Temporal LSTM can consider more contextual information. The structure of Temporal LSTM is shown in Fig. 9.

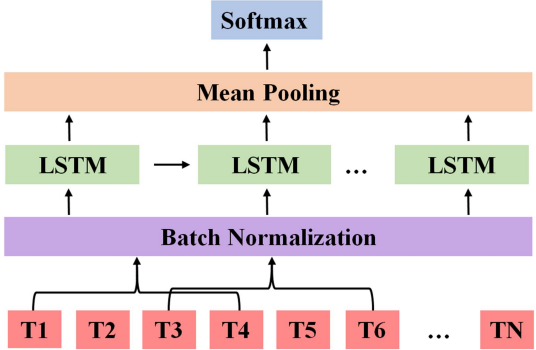

**Figure 9: The structure of Temporal LSTM. Red boxes represent features in different time steps.**

### 3.3 CNN-LSTM

CNN-LSTM is an end-to-end classifier. Mean Pooling LSTM and Temporal LSTM are all multi-step process, where features are extracted first and then fed into classifiers. However, targets of multi-step process are not consistent. Besides, there is no agreement on appropriate features for the emotion classification. To solve these problems, we introduce the end-to-end classifier – CNN-LSTM, whose structure is shown in Fig. 10.

CNN-LSTM takes raw images as inputs. The CNN network is treated as a feature extractor, which extracts the high-level representation for inputs. Then LSTM is utilized to capture temporal information. The whole structure is trained in the end-to-end manner.

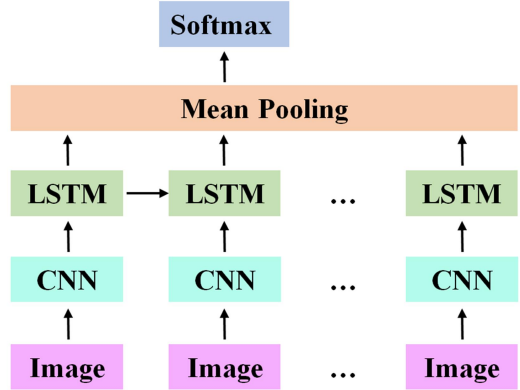

**Figure 10: The structure of CNN-LSTM.**

### 3.4 Aggregation Models

Aggregation models have shown their performance in the Youtube 8M Large-Scale Video Understanding challenge [49]. It is an efficient way to remember all of the relevant visual cues.

We investigate several types of trainable aggregation models, including NetVLAD, NetRVLAD, SoftDBoW and NetFV [32].

As VLAD encoding is not trainable in DNNs, the NetVLAD architecture is proposed to reproduce the VLAD encoding in a trainable manner. Therefore parameters can be optimized through backpropagation instead of using k-means clustering. The NetVLAD descriptor can be written as:

$$NetVLAD(j,k) = \sum_{i=1}^{N} a_k(x_i)(x_i(j) - c_k(j)) \quad (3)$$

$a_k(x_i)$ is the soft assignment of descriptor $x_i$ to cluster $k$. NetVLAD descriptor computes the weighted sum of residuals $(x_i\text{-}c_k)$ of descriptors $x_i$ from learnable anchor point $c_k$ in cluster $k$.

The SoftDBoW and NetFV descriptor exploit the same idea in the NetVLAD descriptor to imitate FV and BOW. Compared with the NetVLAD descriptor, the NetRVLAD descriptor averages the actual descriptors instead of residuals.

## 4 FUSION METHODS

Weighted mean [5, 30] is an efficient late fusion method in previous EmotiW challenges. However, how to efficiently compute weights for a subset of models and ignore useless models are still controversial.

In this paper, we propose the BS-Fusion, which learns from the bream search method. As there is a combinatorial explosion in the number of feasible subset ($2^N$ subsets for $N$ models), we employ a sampling procedure with the goal of filtering out subsets that are less likely to yield good results. We use a beam search of the size $K$ and select $topK$ subsets in each turn. The selection approach is based on the classification accuracy of the subset.

**Algorithm 1** Beam Search Fusion (BS-Fusion)

1: **procedure** BS-Fusion($K$, $N$)
2:    Init empty storage $S$
3:    $S^i$ denotes $i^{th}$ components in $S$
4:    $pre\_best\_score \leftarrow 0$ ; $now\_best\_score \leftarrow 0$
5:    **for** $round = 1,\ldots,N$ **do**
6:      $pre\_best\_score \leftarrow now\_best\_score$
7:      Init empty storage $S^*$
8:      **for** $i = 1,\ldots,K$ **do**
9:        **for** $j = 1,\ldots,N$ **do**
10:          **if** model $j$ not in $S^i$ **then**
11:            $S\_test \leftarrow S^i \cup j$
12:            $S\_test\_score \leftarrow$ calculate scores for $S\_test$
13:            **if** $S\_test\_score > pre\_best\_score$ **then**
14:               $S^* \leftarrow S^* \cup S\_test$
15:      **if** $S^*$ is empty **then**
16:        break
17:      $S \leftarrow topK(S^*)$

## 5 DATASETS

The AFEW database (version 2018) contains video clips labeled using the semi-automatic approach defined in [1]. There are 1809 video clips: 773 for training, 383 for validation and 653 for testing. LBPTOP features and the meta-data are also provided for the Training dataset and the Validation dataset. Category distribution of the AFEW dataset is shown in Table 1.

**Table 1: Emotion Category Distribution of the AFEW Dataset**

| Emotion | Training | Validation | Testing |
|---|---|---|---|
| Angry | 133 | 64 | 98 |
| Disgust | 74 | 40 | 40 |
| Fear | 81 | 46 | 70 |
| Happy | 150 | 63 | 144 |
| Neutral | 144 | 63 | 193 |
| Sad | 117 | 61 | 80 |
| Surprise | 74 | 46 | 28 |
| Total | 773 | 383 | 653 |

## 6 EXPERIMENTAL RESULTS

In this section, we investigate the performance of audio, video and text features. Furthermore, we demonstrate the effectiveness of the BS-Fusion.

### 6.1 Audio Feature Analysis

Since statistical functions have been considered, the feature dimensions of utterance-level features are fixed. We only evaluate their performance in SVM, RF and LR.

Feature dimensions of segment-level features and frame-level features are variable due to variable-length waveforms. Since classifiers take fixed-length features as inputs, we test two methods to compress variable-length features into fixed-length features. As for statistical functions, mean, maximum and FV are utilized to extract fixed-length features. Then we pass them into classifiers such as SVM, RF and LR. As for aggregation models and temporal models, variable-length features are padded into fixed-length features. Then aggregation models (such as NetFV, NetVLAD, NetRVLAD and SoftDBoW) and temporal models (such as Mean Pooling LSTM, Temporal LSTM and CNN-LSTM) are tested.

Through experimental analysis, we find that FV has the worst performance among statistical functions. Although CNN-LSTM gains highest accuracy on the training dataset compared with Mean Pooling LSTM and Temporal LSTM, it has the overfitting problem in the validation dataset. Temporal LSTM gains similar results compared with Mean Pooling LSTM. Therefore, FV is ignored and LSTM refers to Mean Pooling LSTM in the following experiments.

*6.1.1 Results of Temporal Models and Aggregation Models.* In this section, we compare the performance of LSTM, NetVLAD, NetRVLAD, SoftDBoW and NetFV. Experimental results are listed in Table 2.

In the experiments, we choose segment-level audio features, including SoundNet Bottleneck features, MFCCs, IS10 and eGemaps. Segment length for SoundNet Bottleneck features is set to be 1000ms, and segment length for other features is set to be 100ms. The number of neurons in the LSTM layer and the number of neurons in the fully-connected layer are fixed as 128. The cluster size of NetVLAD, NetRVLAD, SoftDBoW and NetFV is set to be 64.

**Table 2: Comparison of Temporal Models and Aggregation Models for Audio Features (%)**

|  | 1000ms SoundNet | 100ms MFCCs | 100ms IS10 | 100ms eGemaps |
|---|---|---|---|---|
| NetVLAD | 32.64 | 26.63 | 21.41 | **27.94** |
| NetRVLAD | 33.68 | 24.80 | 19.58 | 26.11 |
| NetFV | 32.11 | **27.68** | 21.41 | 26.11 |
| SoftDBoW | 32.38 | 25.85 | 20.37 | 27.42 |
| LSTM | **34.99** | 27.15 | **24.03** | 26.11 |

Through experimental results in Table 2, we find that LSTM has better performance in most cases. Therefore, we only consider LSTM in the following experiments and ignore aggregation models.

*6.1.2 Performance of Audio Features.* In this section, we compare the performance of multiple audio features. Experimental results are listed in Table 3.

**Table 3: Classification Accuracy of Audio Features (%)**

| Exp. | Features | Statistical Functions | Classifiers | Accuracy |
|---|---|---|---|---|
| 1 | 1000ms SoundNet | None | LSTM | 31.33 |
| 2 | 1000ms VGGish | None | LSTM | 34.86 |
| 3 | 1000ms Chinese ASR | Max | RF | **36.03** |
| 4 | 1000ms English ASR | Mean | RF | 33.42 |
| 5 | 100ms eGemaps | None | LSTM | 26.89 |
| 6 | 100ms IS10 | Max | RF | 25.59 |
| 7 | 100ms MFCCs | None | LSTM | 26.63 |
| 8 | eGemaps | — | RF | 34.46 |
| 9 | IS09 | — | RF | 32.11 |
| 10 | IS11 | — | RF | 21.15 |
| 11 | IS13 | — | RF | 20.10 |

Exp. 1~7 in Table 3 choose segment-level audio features. Exp. 8~11 in Table 3 test multiple utterance-level audio features. As for segment-level features, we list segment length in front of the feature name. As statistical functions are not needed for LSTM, they are set to be *None*.

Through experimental results in Table 3, we find that different audio features need different statistical functions and classifiers. Chinese ASR bottleneck features gain the highest accuracy, 36.03%. As the Chinese ASR system trains on a larger speech corpus than the English ASR system, Chinese ASR bottleneck features have better performance. It shows the efficiency of features extracted from the multilingual system.

## 6.2 Video Feature Analysis

In this section, we show our face detection approach and the performance of video features.

*6.2.1 Face Detection Methods.* In provided faces, 17 videos in the training dataset and 12 videos in the validation dataset are false detected. As for false detected videos, we manually initial the position of the first face and then use the object tracking method to extract the following faces. In the end, we convert faces into grey-scale images and apply histogram equalization to alleviate the impact of lights.

*6.2.2 Performance of Video Features.* We extract bottleneck features from both SFEW fine-tuned models and FER+ fine-tuned models. We find that SFEW fine-tuned models gain worse performance compared with FER+ fine-tuned models. Therefore, only FER+ fine-tuned models are considered.

**Table 4: Classification Accuracy of Video Features (%)**

| Exp. | Features | Statistical Functions | Classifiers | Accuracy |
|---|---|---|---|---|
| 1 | DenseNet_pool3 | None | LSTM | 41.25 |
| 2 | VGG_conv5-b | Mean | RF | 43.08 |
| 3 | VGG_conv5-c | None | LSTM | **43.34** |
| 4 | VGG_fc1 | Mean | RF | 41.78 |
| 5 | VGG_fc2 | None | LSTM | 39.16 |
| 6 | C3D_fc6 | None | LSTM | 37.86 |
| 7 | Geo-Features | Max | SVM | 28.86 |
| 8 | Background | None | LSTM | 24.54 |
| 9 | Identity Features | Mean | RF | 36.81 |
| 10 | LBPTOP | — | SVM | 38.81 |
| 11 | LBP | — | SVM | 29.24 |
| 12 | HOG | — | RF | 39.95 |
| 13 | HOGLBP | — | RF | 40.73 |
| 14 | DSIFT | — | RF | 39.69 |

Exp. 1~9 in Table 4 choose frame-level features or segment-level features. Exp. 10~14 in Table 4 evaluate multiple video-level features. Through experimental results, we find that different video features need different statistical functions and classifiers. VGG_conv5-c features gain the highest accuracy, 43.34%, which outperform the best result in the audio modality. HOGLBP features are the best handcraft features, which gains 40.73% accuracy. Through Exp. 7~9 in Table 4, we find that our newly proposed features have worse performance compared with other visual features. However, through further experiments, we find that those features (especially Identity features) are helpful during the fusion phase. We can gain higher accuracy if we take into account those features.

## 6.3 Text Feature Analysis

We utilize the open-source Baidu API[1] to recognize audio contents. To reduce the size of the vocabulary, we remove the word whose frequency is less than three. Furthermore, we change the word to its prototype. For example, 'go', 'going' and 'gone' are all converted into 'go'. Then TF-IDF and WV features are extracted.

**Table 5: Classification Accuracy of Text Features (%)**

| Exp. | Features | Statistical Functions | Classifiers | Accuracy |
|---|---|---|---|---|
| 1 | WV | Max | SVM | **36.94** |
| 2 | TF-IDF | — | RF | 27.68 |

Through experimental results in Table 5, we find that WV features are more suitable for the limited dataset. WV features gain the highest accuracy, 36.94%, which outperform the best features in the audio modality. It shows the effectiveness of textual features.

## 6.4 Fusion Results

Through the BS-Fusion, a subset of emotion possibilities is selected according to the classification performance on the validation dataset. In the testing dataset, we achieve 60.34% accuracy.

---

[1]Baidu API: http://ai.baidu.com/docs#/ASR-Online-Python-SDK/top

Figure 11: The confusion matrix in the testing dataset.

Through Fig. 11, we can figure out that our approach has great recognition performance in *angry*, *happy* and *neutral*. However, *disgust* and *surprise* are easily confused with others.

*Surprise* is easily confused with *fear* and *neutral*. It is related to the vague definition of *surprise*. *Surprise* contains pleasant surprises and fright. Pleasant surprises are easy confused with *happy*. And fright is easily confused with *fear*.

*Disgust* is more blurred than *surprise*. *Disgust* is related to the video content. If we add the video description information, the recognition accuracy of *disgust* can be increased.

## 7 CONCLUSIONS

In this paper, we present the audio-video-text based emotion recognition system submitted to EmotiW 2018. Features from different modalities are trained individually. Then emotion possibilities are extracted and passed into the BS-Fusion. We evaluate our method in the EmotiW 2018 Audio-Video based sub-challenge. Multiple features and classifiers are investigated. Through experimental analysis, we find that the video modality has the highest recognition accuracy among three modalities. Finally, we achieve 60.34% recognition accuracy in the testing dataset via the BS-Fusion.

In the future, we will add more discriminative features for emotion recognition. Since emotion expression is related to the video content, video description information will be considered. Furthermore, movie types also count. *Fear* is common in horror films.